\documentclass{article}

\usepackage{microtype}
\usepackage{graphicx}
\usepackage{booktabs} 

\usepackage{hyperref}

\usepackage[accepted]{synsml2023}

\usepackage{caption}
\usepackage{subcaption}

\usepackage{amsmath}
\usepackage{amssymb}
\usepackage{mathtools}
\usepackage{amsthm}

\usepackage[capitalize,noabbrev]{cleveref}

\theoremstyle{plain}

\theoremstyle{definition}

\theoremstyle{remark}

\usepackage[textsize=tiny]{todonotes}

\synsmltitlerunning{Learning Green's Function Efficiently Using Low-Rank Approximations}

\begin{document}

\twocolumn[
\synsmltitle{Learning Green's Function Efficiently Using Low-Rank Approximations}


\synsmlsetsymbol{equal}{*}

\begin{synsmlauthorlist}
\synsmlauthor{Kishan Wimalawarne}{utokyo}
\synsmlauthor{Taiji Suzuki}{utokyo,aip}
\synsmlauthor{Sophie Langer}{twente}
\end{synsmlauthorlist}

\synsmlaffiliation{utokyo}{Department of Mathematical Informatics, The University of Tokyo, Tokyo, Japan}
\synsmlaffiliation{aip}{Center for Advanced Intelligence Project (AIP), RIKEN, Tokyo, Japan}
\synsmlaffiliation{twente}{Faculty of Electrical Engineering, Mathematics, and Computer Science, University of Twente, Enschede, The Netherlands}

\synsmlcorrespondingauthor{Kishan Wimalawarne}{kishanwn@gmail.com}

\synsmlkeywords{Machine Learning}

\vskip 0.3in
]



\printAffiliationsAndNotice{\synsmlEqualContribution} 

\begin{abstract}
Learning the  Green's function using deep learning models enables  to  solve  different classes of partial differential equations. 
A practical limitation of using deep learning for the Green's function is the repeated computationally expensive Monte-Carlo integral  approximations. We propose to learn the Green's function by low-rank decomposition, which results in a novel architecture to remove redundant computations by separate learning with  domain data for evaluation and Monte-Carlo samples for integral approximation. Using experiments we show that the proposed method  improves computational time compared to MOD-Net while achieving  comparable accuracy compared to both PINNs and MOD-Net.
 \end{abstract}

\section{Introduction}
The Green's function is a well-known method to solve and analyze Partial Differential Equations (PDE) \cite{evans10,Bebendorf_Hackbusch}. Recently, deep learning models have been applied to  learn the Green's function enabling to obtain solutions for both linear and nonlinear PDEs \cite{modnet} as well as to learn  PDEs in irregular domains \cite{msml-teng22a}. 
\citet{modnet} further demonstrated that deep learning models can parameterize a class of PDEs compared to learning a specific PDE. 

The MOD-Net \cite{modnet} uses the Green's function approximation by using a neural network and Monte-Carlo integration to parameterize solutions for PDEs. A limitation of the MOD-Net is that the approximation of the Green's function by Monte-Carlo integration leads to high computational costs due to their repeated evaluations for each domain element. In this research, we propose to extend MOD-Net with low-rank decomposition of the  Green's function. The proposed model results in a two network architecture, where one network learns on the domain elements to be evaluated and other network learns over all the Monte-Carlo samples, hence, avoid redundant repeated computations of Green's function at each domain element. 
Using experiments with the Poisson 2D equation and the linear reaction-diffusion equation we show that our proposed method is  computationally feasible compared to MOD-Net, while achieving comparable accuracy compared to PINNs \cite{PINNS_1} and MOD-Net. Additionally, we show that our proposed method has the ability to interpolate within the solution space as a neural operator similar to MOD-Net. 

\section{Learning Green's Function by MOD-Net} 
Let us  consider a domain $\Omega \subset \mathbb{R}^{d}$, a differential operator $\mathcal{L}$, a source function $g(\cdot)$ and a boundary condition $\phi(\cdot)$,  then a   partial differential equation can be represented as,
\begin{align*}
\mathcal{L}[u](x)  = g(x), \qquad& x \in \Omega\\ \nonumber
u(x) = \phi(x), \qquad& x \in \partial\Omega. 
\end{align*}

For a linear PDE with the Dirichlet boundary condition $\phi(\cdot) =0$, we can find a Green's function $G:\mathbb{R}^{d} \times \mathbb{R}^{d} \rightarrow \mathbb{R}$ for a fixed $x' \in \Omega$ as follows: 
\begin{align*}
\mathcal{L}[G](x) & = \delta(x -x'), \qquad x \in \Omega\\ \nonumber
G(x,x') &= 0, \qquad \qquad \quad\; x \in \partial\Omega, \label{eq:green_delta_interior}
\end{align*}
which leads to a solution function $u(\cdot)$ as
\begin{equation}
u(x) = \int_{\Omega} G(x,x')g(x')dx'. \label{eq:green_formulation_general}
\end{equation}

Recently developed MOD-Net \cite{modnet}  proposes to learn the Green's function by using a neural network. It uses a neural network $G_{\theta_1}(x,x')$ with parameters denoted by $\theta_1$ to replace the operator  $G(x,x')$ of \eqref{eq:green_formulation_general}.  
Given a set $S_{\Omega}$ consisting of random samples from $\Omega$, the Monte-Carlo approximation of the  Green's function integral  \eqref{eq:green_formulation_general} results in the following:
\begin{equation}
u_{\theta_1}(x;g) = \frac{|\Omega|}{|S_{\Omega}|} \sum_{x' \in S_{\Omega}} G_{\theta_1}(x,x')g(x'). \label{eq:monte_carlo_u}
\end{equation}
MOD-Net \cite{modnet} also proposes of a nonlinear extension of the above as 
\begin{equation}
u_{\theta_1,\theta_2}(x;g) = F_{\theta_2} \left(
\frac{|\Omega|}{|S_{\Omega}|} \sum_{x' \in S_{\Omega}} G_{\theta_1}(x,x')g(x') \right), \label{eq:nonliear_modnet}
\end{equation}
where $F_{\theta_2}$ is a neural network ($\theta_2$ representing parameters). 

It has been shown  \citep{modnet} that Green's function can learn as an operator for varying $g(\cdot)$ to parameterize  a  class of PDE.
Following \citep{modnet}, let us consider $K$ different parametarizations of a PDE class by specifying  $g^k(\cdot)$ for $k=1,\cdots,K$.
Let $S^{\Omega,k}$ and $S^{\partial\Omega,k}$ are domain elements from the interior and boundary, respectively, for each $k=1,\ldots,K$. Then the objective function for MOD-Net for the formulation in \eqref{eq:monte_carlo_u}  is given as
\begin{multline}
R_{S}   = \\
\frac{1}{K} \sum_{k \in [K]}\Bigg( 
\lambda_1 \frac{1}{|S^{\Omega,k}|} \sum_{x \in S^{\Omega,k}}\|\mathcal{L}[u_{\theta_1}(x;g^k)](x) - g^k(x) \|_2^2 \\
+ \lambda_2 \frac{1}{|S^{\partial\Omega,k}|} \sum_{x \in S^{\partial\Omega, k}}\|u_{\theta_1}(x;g^k) \|_2^2 
 \Bigg), \label{eq:loss_modnet}
\end{multline}
where $\lambda_1$ and $\lambda_2$ are regularization parameters.
Notice that when $K=1$ the above objective function learns  a specific instance of a PDE. 

A major limitation with MOD-Net is the computational bottleneck due to the Monte-Carlo approximation of the integrals \eqref{eq:monte_carlo_u} and \eqref{eq:nonliear_modnet}. The repeated summation by elements of $S_{\Omega}$ with respect to each domain element in \eqref{eq:loss_modnet} can become both computationally costly and redundant. 

\section{Proposed Method}
We propose to extend MOD-Net with low-rank presentation of the Green's function to remove redundant computations and improve computational feasibility. 

First, we put forward the following approximation for low-rank decomposition of the Green's function from \citet{Bebendorf_Hackbusch}.
For any $0<\epsilon<1$ sufficiently small and $R \geq c^d \lceil\log(\epsilon^{-1})\rceil^d + \lceil\log(\epsilon^{-1})\rceil$, and  $D_1,D_2 \subset \Omega$, \citet{Bebendorf_Hackbusch} have given decomposition with functions $v_i(\cdot)$ and $w_i(\cdot)$, $i=1,\ldots,R$ as 
\begin{equation}
G_R(x,y) = \sum_{i=1}^{R} v_i(x)w_i(y),\;\; x\in D_1,\;y\in D_2, \label{eq:low_rank_dec} 
\end{equation}
such that 
\begin{equation}
\|G(x,\cdot) - G_R(x,\cdot)\|_{L^{2}(D_1)} \leq \epsilon \|G(x,\cdot) \|_{L^{2}(\hat{D}_1)},\label{eq:low_rank_dec_1} 
\end{equation}
where $\hat{D}_1 \subset \Omega$ a set slightly large than $D_1$.

We now apply the above low-rank decomposition \eqref{eq:low_rank_dec} of the Green's function to \eqref{eq:monte_carlo_u}. We propose to learn functions $u_i(x)$ and $v_i(x), \; i=1,\ldots,R$ using two neural networks $F_{\gamma_1}:\mathbb{R}^{d} \rightarrow \mathbb{R}^{R}$ and $H_{\gamma_2}:\mathbb{R}^{d} \rightarrow \mathbb{R}^{R}$ where $\gamma_1$ and $\gamma_2$ represent parameters. In practice, $R$ can be problem and data dependent, hence, may require to be considered as a hyperparameter. Further, we assume that $D_1=D_2=\hat{D}_1=\Omega$ for \eqref{eq:low_rank_dec} and \eqref{eq:low_rank_dec_1}  and there exist some neural network that can learn a low-rank representation. 

Next, we apply the learning of low-rank decomposition to \eqref{eq:monte_carlo_u} which leads to the following expansion 
\begin{align} 
u_{\gamma_1,\gamma_2}(x;g) &= \frac{|\Omega|}{|S_{\Omega}|} \sum_{y \in S_{\Omega}} G(x,y) g(y) \nonumber \\ 
     &\approx  \frac{|\Omega|}{|S_{\Omega}|} \sum_{y \in  S_{\Omega}} \sum_{i=1}^{R} F_{\gamma_1}(x)_iH_{\gamma_2}(y)_i g(y)  \nonumber \\
	&= \frac{|\Omega|}{|S_{\Omega}|} \sum_{i=1}^{R} F_{\gamma_1}(x)_i  \sum_{y \in S_{\Omega}} H_{\gamma_2}(y)_i g(y) \nonumber\\
 &= \frac{|\Omega|}{|S_{\Omega}|} F_{\gamma_1}(x)^{\top} \left[ \sum_{y \in S_{\Omega}} H_{\gamma_2}(y) g(y)\right]. \label{eq:monte_carlo_u_lr}
\end{align}
The last step of \eqref{eq:monte_carlo_u_lr} shows the separation of the learning with Monte-Carlo samples from the set $S_{\Omega}$ by $H_{\gamma_2}(\cdot)$ and learning with each $x$  by $F_{\gamma_1}(\cdot)$. 
This separated learning helps to avoid the computationally costly repeated evaluations in  \eqref{eq:monte_carlo_u} and \eqref{eq:nonliear_modnet} since  we only need to  compute the network $H_{\gamma_2}(\cdot)$ once for all input $x$ at each learning iteration.   

\begin{figure*}[t]
\centering
\begin{subfigure}{.32\textwidth}
  \centering
  \includegraphics[width=.95\linewidth]{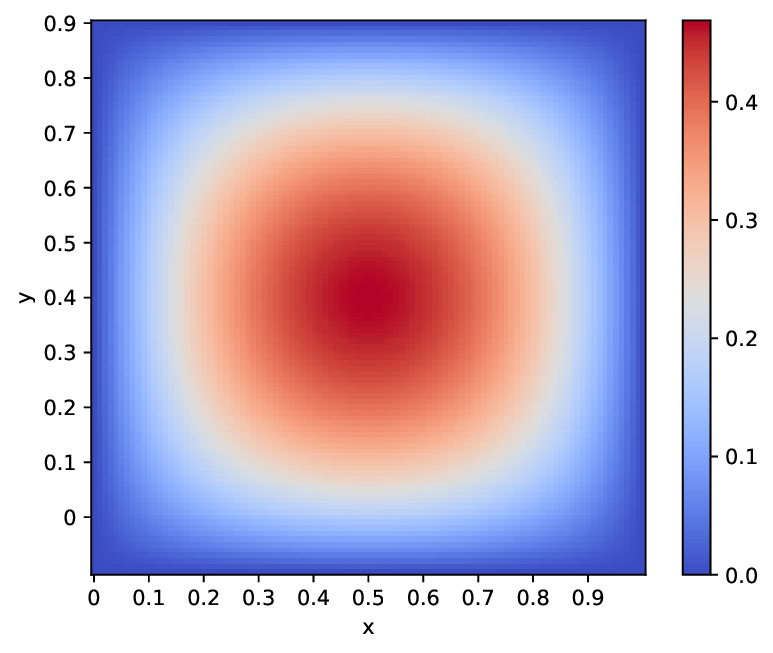}
  \caption{Exact solution}
  \label{fig:sfig1}
\end{subfigure}%
\begin{subfigure}{.32\textwidth}
  \centering
  \includegraphics[width=.95\linewidth]{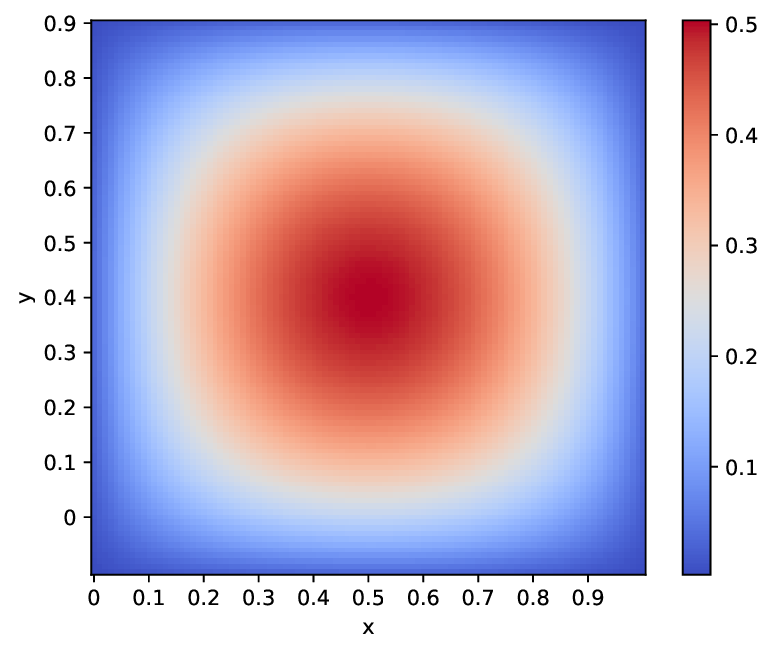}
  \caption{Predicted solution}
  \label{fig:sfig2}
\end{subfigure}
\begin{subfigure}{.32\textwidth}
  \centering
  \includegraphics[width=.95\linewidth]{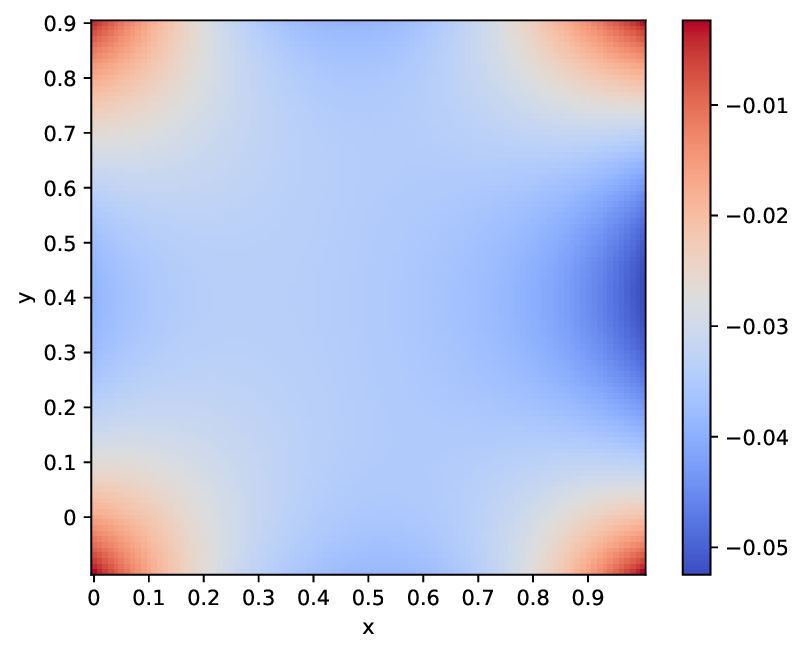}
  \caption{Error}
  \label{fig:sfig2}
\end{subfigure}
\label{fig:linear_poisson_2d_15}
\caption{Interpolation results for Poisson 2D problem by DecGreenNet-NL for the parameter $a=15$ (a) Exact solution, (b) Predicted solution by DecGreenNet-NL and (c) Error}
\end{figure*}

\begin{table*}[t]
\centering
\begin{tabular}{|l|l|l|l|l|} \hline
Method & Network structure & P & Loss & Time(sec)   \\ \hline 
PINNs &[2, 64, 64, 1] & - & 3.5e-4 &  80.14  \\ \hline
DecGreenNet &$F_{\gamma_{1}}$ = [2,512, 512, 512, 512, 50],$H_{\gamma_{2}}$ = [2, 16, 16, 16, 1] & 100 & 2.86e-4 & 117.95 \\  \hline
DecGreenNet-NL &$F_{\gamma_{1}}$ = [2,64,64,64,64,64,50],$H_{\gamma_{2}}$ = [2,64,64,64,50],$O_{\gamma_{3}}$ = [100,1] & 100 & 1.5e-3 & 550.45   \\ \hline
MOD-Net &[4, 128, 128, 128, 128, 1]  & 10 & 1.05e-3 & 721330\\ \hline
\end{tabular}
\caption{Network structures, number of random samples, test loss, and computational times for a single instance learning of Poisson 2D equation  for $a=15$ using PINNs, DecGreenNet, DecGreenNet-NL, and MOD-Net} \label{tb:poi2d_15_table}
\end{table*}

Using the construction in \eqref{eq:monte_carlo_u_lr} we propose two neural network architectures to improve over \eqref{eq:monte_carlo_u} and \eqref{eq:nonliear_modnet}. For simplicity, we omit the factor $|\Omega|/|S_{\Omega}|$ due to its redundancy in the learning process. Further, we assume that elements $S_{\Omega}$ are sampled once and fixed during the learning and prediction processes. 

We put forward  \textit{DecGreenNet} as a direct construction from  \eqref{eq:monte_carlo_u_lr} with $|S_{\Omega}|=P$ as  
\begin{equation} 
u_{\gamma_1,\gamma_2}(x;g) =  F_{\gamma_1}(x)^{\top}
\sum_{i=1}^{P} H_{\gamma_2}(y_i) g(y_i). \label{eq:DecGreenNet_u}
\end{equation}
Next, we  construct a nonlinear extension of MOD-Net \eqref{eq:nonliear_modnet} with the low-rank  decomposition following \eqref{eq:monte_carlo_u_lr}. Here we remove the summation on the right and arrange its elements as  a concatenation. By introducing an additional neural network $O_{\gamma_3}:\mathbb{R}^{P} \rightarrow \mathbb{R}$, we propose  \textit{DecGreenNet-NL} as  
\begin{multline} 
u_{\gamma_1,\gamma_2,\gamma_3}(x;g) =  O_{\gamma_3}\bigg(F_{\gamma_1}(x)^{\top}\mathrm{concat}\big[H_{\gamma_2}(y_1) g(y_1), \\
H_{\gamma_2}(y_2) g(y_2), \ldots, H_{\gamma_2}(y_P) g(y_P) \big]\bigg), \label{eq:DecGreenNet_nl_u}
\end{multline}
where $\mathrm{concat}(\cdots)$ is an operation to concatenation of input vectors.
Note that when $O_{\gamma_3}$ is replaces by a vector of ones ($O_{\gamma_3} = [1,1,\ldots,1] \in \mathbb{R}^{P}$), \eqref{eq:DecGreenNet_nl_u} becomes equivalent to \eqref{eq:DecGreenNet_u}.

Optimal neural architectures (layers and hidden units) of $F_{\gamma_1}$, $H_{\gamma_1}$, and $O_{\gamma_3}$ need to be discovered by hyperparameter tuning. Additionally, $R$ and $P$ should be considered as hyperparameters. Furthermore, the activation functions for neural networks require higher-order differential capacity in relation to the differential operators in the PDE. In general, we use the activation function $\mathrm{ReLUK}(x) :=\max\{0,x\}^K$ where $K \in \mathbb{N}_{+}$.

\begin{figure*}[t]
  \centering
\begin{subfigure}{.32\textwidth}
  \centering
  \includegraphics[width=0.95\linewidth]{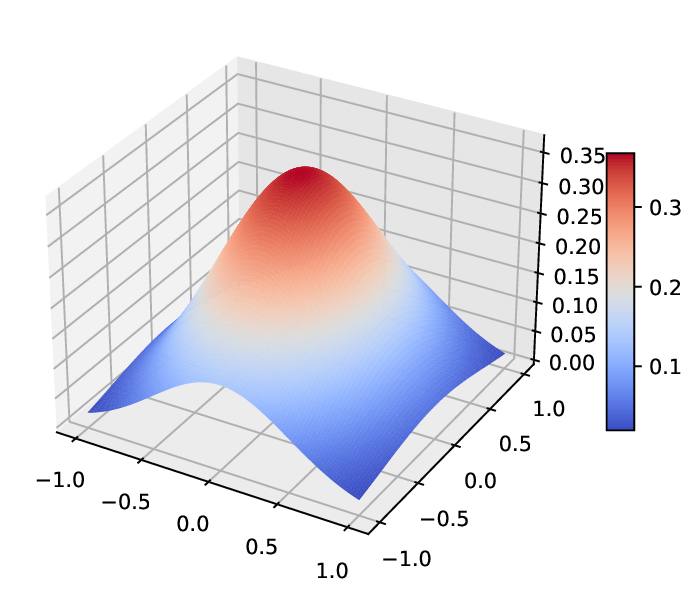}
  \caption{Exact solution}
  \label{fig:sfig1}
\end{subfigure}%
\begin{subfigure}{.32\textwidth}
  \centering
  \includegraphics[width=.95\linewidth]{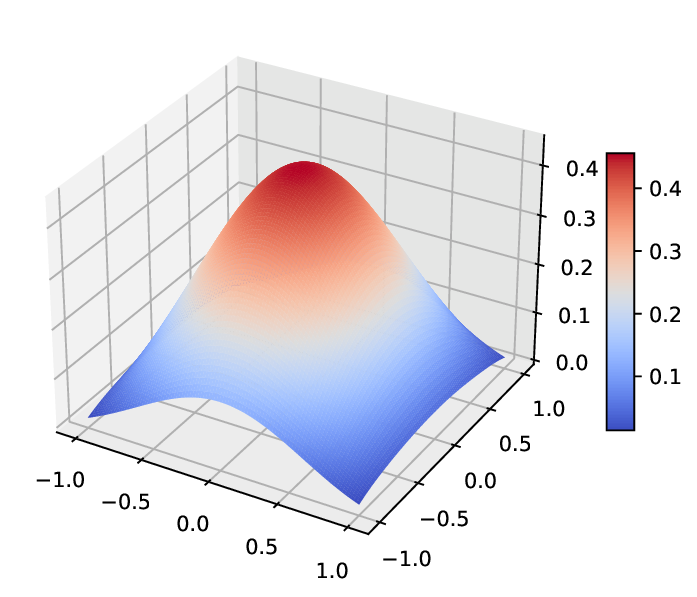}
  \caption{Predicted solution}
  \label{fig:sfig2}
\end{subfigure}
\begin{subfigure}{.32\textwidth}
  \centering
  \includegraphics[width=.85\linewidth]{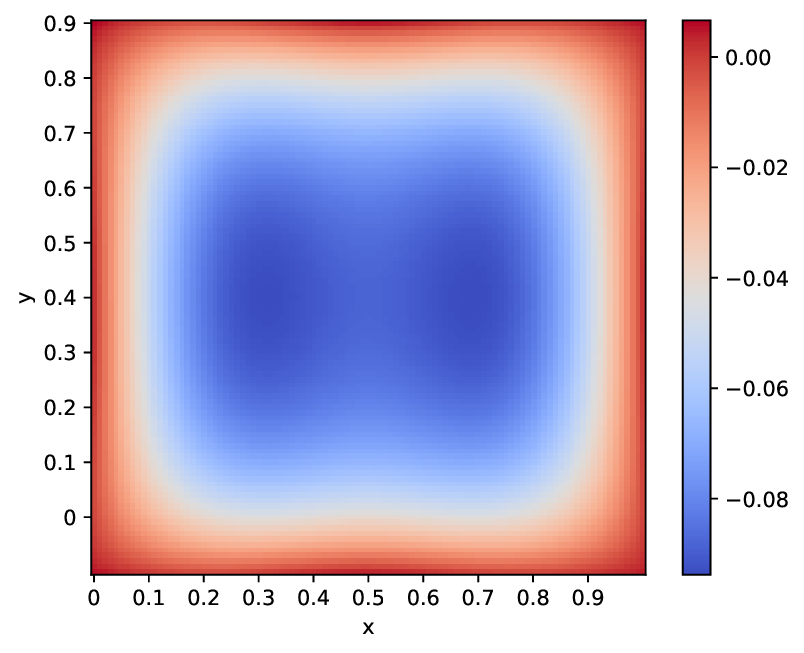}
  \caption{Error}
\end{subfigure}
\label{fig:linear_rf} 
\caption{Comparison between the exact solution and predicted solution by DecGreenNet of the reaction-diffusion equation  (a) Exact solution, (b) Predicted solution by DecGreenNet and (c) Error.} 
\end{figure*}

\begin{table*}[t]
\centering
\begin{tabular}{|l|l|l|l|l|} \hline
Method & Network structure & P &  Loss  & Time (sec)   \\ \hline 
PINNs & [2,128, 128, 128, 128, 128,1] & - & 2.6e-4 &  99.48  \\ \hline
DecGreenNet & $F_{\gamma_{1}}$ = [2,512, 512, 512, 512, 512,50], $H_{\gamma_{2}}$ = [2,32,32,32,50] & 300 & 7.76e-5 &  355.25  \\ \hline
DecGreenNet-NL & $F_{\gamma_{1}}$ = [2, 256, 256, 256, 256, 256, 50], $H_{\gamma_{2}}$ = [2, 64, 64, 64, 64, 50] & 300 & 2.70e-4 &  1075.24  \\
 &  $O_{\gamma_{3}}$ = [300,1] &  &  &    \\ \hline
MOD-Net &  [4,128,128,128,1]  &  10  & 1.26e-4  & 23696   \\ \hline 
\end{tabular}
\caption{Network structures, number of random samples, test loss, and computational times for the linear reaction-diffusion equation using PINNs, DecGreenNet, DecGreenNet-NL, and MOD-Net} \label{tb:linear_rf}
\end{table*}

\section{Experiments}
We experimented with PDEs used in \cite{modnet} and \cite{msml-teng22a} to evaluate our proposed models.  

\subsection{Experimental Setup}
In all our experiments we set $\lambda_1=\lambda_2=1$ in the objective function \eqref{eq:loss_modnet} for all models. For both DecGreenNet and DecGreenNet-NL selected the layers and hidden units of each neural network by hyperparameter tuning. We selected layers from $1,\ldots,6$ and hidden units from  $2^h\; h =3,\ldots,6$ for both $F_{\gamma_1}(\cdot)$ and $H_{\gamma_2}(\cdot)$ of \eqref{eq:DecGreenNet_u} and \eqref{eq:DecGreenNet_nl_u}. For $O_{\gamma_3}(\cdot)$ we selected from hidden layers $0,1,2$ and hidden units $4,8,16$. We represent the network structure of network of models by notation $[{in},h,\ldots,h,{out}]$, ${in}$, ${out}$, and $h$ represent dimensions of input, output, and hidden layers, respectively. We experimented with $R \in \{5,50,100\}$. We used PINNs as a baseline method by performing hyperparameter tuning for layers and hidden units varying from $1,\ldots,5$ and $2^h\; h =3,\ldots,6$, respectively. We also used MOD-Net as a baseline method, however,  we only used $10$ random samples to approximate the Green's function due to high computational cost. We used the activation function $\mathrm{ReLU3}(\cdot)$ for all models. We used the Pytorch environment with Adam optimization method with learning rate of $0.001$ and weight decay set to zero. All  experiments were conducted on NVIDIA A100 GPUs with CUDA 11.6 on a Linux 5.4.0 environment. We provide the code at \url{https://github.com/kishanwn/DecGreenNet}.

\subsection{Poisson 2D Equation}
As our first experiment, we used the Poisson 2D problem with multiple paramterizations under the same setting used in \cite{modnet}. The  Poisson 2D problem for the  the domain of $\Omega = [0,1]^2$ is specified  as
\begin{align*}
-\Delta u(x,y) &= g(x,y), \;\;(x,y) \in \Omega,\\
u(x,y) &= 0, \qquad\quad (x,y) \in \partial\Omega. 
\end{align*}
where $g(x,y) = -a(x^2 - x + y^2 -y)$. 
The analytical solution of the above problem is $u(x,y) = \frac{a}{2}x(x-1)y(y-1)$.
In \cite{modnet}, multiple parameterizations of the Poisson equation is specified by setting different values for $a$.  Following a similar setting as in \cite{modnet}, we used $a:=a_k=10k$ where $k=1,\cdots,10$ which lead to $g(x,y) := g^{k}(x,y) = -a_k(x^2 - x + y^2 -y),\; k=1\dots,10$.  

We found that  DecGreenNet-NL with  $F_{\gamma_1}=[2, 256, 256, 256, 256, 256, 50]$, $H_{\gamma_2}=[2, 64, 64, 64, 64, 50]$, and $O_{\gamma_3} = [100,1]$ provided the best solution. From the learned model, we interpolate the solution for  the Poisson 2D equation  with $a=15$. The low error of the interpolated solution in Figure 1 shows the  operator learning capability of our method, hence, the ability to learn parameterization for a class of PDE.


As our second experiment with Poisson 2D, we conducted experiments to evaluate solutions for a single instance of the Poisson 2D equation with $a=15$ ($K=1$ in \eqref{eq:loss_modnet}).  Table \ref{tb:poi2d_15_table} shows the details on learning with PINNs, MOD-Net, and single instance learning by DecGreenNet and DecGreenNet-NL. Both our proposed methods obtained  lower values for test loss  compared to PINNs and MOD-Net, in addition to the significantly small time compared to MOD-Net.



\subsection{Reaction-Diffusion Equation}
We  experimented with the linear reaction-diffusion equation \cite{msml-teng22a} in the domain $\Omega \in [-1,1]^2$ specified as 
\begin{equation}
\mathcal{L}(u) = -\nabla \cdot ((1 +2x^2)\nabla) + (1+y^2)u 
\end{equation}
The exact solution is $u(x,y) = e^{-(x^2 + 2 y^2 +1)}$.
From both Table \ref{tb:linear_rf} and Figure 2, we observe that DecGreenNet obtains a good accuracy for learning the linear reaction-diffusion function with moderate computational time. 

\section{Broader Impact}
 We believe that the computational advantages and operator learning capability of our proposed method would be conducive to solving important problems in science. A limitation of our method is the considerable amount of hyperparameters  that needed to be tuned  to find the  optimal neural architectures. We do not know any negative impacts from our proposed methods.

\section{Acknowledgement}
TS was partially supported by JSPS KAKENHI (20H00576) and JST CREST. KW was partially supported by JST CREST.

\section{Conclusion and Future Work}
We provide a computationally feasible low-rank model to learning PDEs with Green's function. Theoretical analysis  such as convergence bounds for our model is  open for future work. Further extensions of our model to solve high-dimensional PDEs is another future direction.  

\bibliographystyle{synsml2023} 
{\small
\bibliography{ref}
}

\end{document}